\documentclass{article}

     \usepackage[nonatbib,preprint]{neurips_2020}

\usepackage[utf8]{inputenc} 
\usepackage[T1]{fontenc}    
\usepackage{hyperref}       
\usepackage{url}            
\usepackage{booktabs}       
\usepackage{amsfonts}       
\usepackage{nicefrac}       
\usepackage{microtype}      
\usepackage{slashbox}
\usepackage{amsmath}
\usepackage{ dsfont }
\usepackage{graphicx}
\usepackage{subcaption}
\usepackage[backend=biber,style=numeric-comp,sorting=nty]{biblatex}
\usepackage{authblk}

\title{Robust Semantic Interpretability: Revisiting Concept Activation Vectors}

\author[1,6]{\textbf{Jacob Pfau}}
\author[1,6]{\textbf{Albert T. Young}}
\author[5,6]{\textbf{Jerome Wei}}
\author[1,2,6]{\textbf{Maria L. Wei}}
\author[3,4,6]{\textbf{Michael J. Keiser}}

\affil[1]{\footnotesize Department of Dermatology, UCSF and Dermatology Service, SFVAMC}
\affil[2]{\footnotesize Helen Diller Family Comprehensive Cancer Center, UCSF}
\affil[3]{\footnotesize Institute for Neurodegenerative Diseases, and Bakar Computational Health Sciences Institute, UCSF}
\affil[4]{\footnotesize Department of Pharmaceutical Chemistry, Department of Bioengineering and Therapeutic Sciences, UCSF}
\affil[5]{\footnotesize Department of Computer Science, UC Berkeley}
\affil[6]{\footnotesize \{pfau, keiser\}@keiserlab.org, \{albert.young, maria.wei\}@ucsf.edu, jeromew@berkeley.edu}

\begin{document}

\maketitle

\begin{abstract}
Interpretability methods for image classification assess model trustworthiness by attempting to expose whether the model is systematically biased or attending to the same cues as a human would. Saliency methods for feature attribution dominate the interpretability literature, but these methods do not address semantic concepts such as the textures, colors, or genders of objects within an image. Our proposed Robust Concept Activation Vectors (RCAV) quantifies the effects of semantic concepts on individual model predictions and on model behavior as a whole. RCAV calculates a concept gradient and takes a gradient ascent step to assess model sensitivity to the given concept. By generalizing previous work on concept activation vectors to account for model non-linearity, and by introducing stricter hypothesis testing, we show that RCAV yields interpretations which are both more accurate at the image level and robust at the dataset level. RCAV, like saliency methods, supports the interpretation of individual predictions. To evaluate the practical use of interpretability methods as debugging tools, and the scientific use of interpretability methods for identifying inductive biases (e.g. texture over shape), we construct two datasets and accompanying metrics for realistic benchmarking of semantic interpretability methods. Our benchmarks expose the importance of counterfactual augmentation and negative controls for quantifying the practical usability of interpretability methods.\footnote{RCAV and dataset code available at https://github.com/keiserlab/rcav.}
\end{abstract}

\section{Introduction}
\label{intro}

Model interpretability is essential to both the development and deployment of safe and robust models. In the development phase, interpretability methods function as debugging tools, verifying whether machine learning models consider the input features which matter most to humans \cite{Ross2017-eo, Holzinger2017-re, Weller2017-ap}. In the deployment phase, interpretability methods check model performance on the fly by showing the user what part of the image the model focuses on \cite{Murdoch2019-qv, Selvaraju2017-ne}.

The most widely used interpretability methods, saliency maps, explain models at the image level by attributing model prediction to input features, usually pixels \cite{Kim2017-yi, Zhang2019-wq, Adel2018-ds}. In contrast, semantic interpretability methods explain models by quantifying concept sensitivity, i.e. the effect of altering the input with respect to one concept -- e.g. color, age, gender, or texture -- while holding constant other aspects of the image. For example, for high-stakes applications such as medical imaging, it is essential to understand whether the model uses reliable semantic concepts (e.g. object shape), or instead relies on the presence of confounding noise (e.g. an image corruption) \cite{Zech2018-ad, Zech2018-qg}. Semantic interpretability methods can uncover bias at the dataset level during model development as well at the image level during deployment. 

In this paper, we propose a state of the art method for quantifying concept sensitivity, and introduce the first realistic benchmarks for semantic interpretability. We consider two use cases of semantic interpretability: (1) image-level quantification of concept sensitivity, e.g. the effect of increasing red-levels in an image, and (2) dataset-level quantification of model concept bias, i.e. the aggregation of image-level concept sensitivity across a dataset. We also identify a gap in the literature: interpretability benchmarks are too artificial and disconnected from these use cases. 

To be useful in practice, a semantic interpretability method must work with existing convolutional neural network (CNN) architectures, minimize added computational and data overhead, and yield quantitative results. To this end, Kim et al. introduced concept activation vectors (CAV) which linearly approximate a model's latent space concept encoding \cite{Kim2017-yi}. Their method, Testing with CAVs (TCAV), quantifies dataset-level concept sensitivity by taking the inner product of a CAV with model output gradients. The proposed Robust Concept Activation Vector (RCAV) method builds on TCAV. Whereas TCAV was restricted to linear interaction between concept and model, we generalize to allow for non-linear effects of concepts on model predictions.

Our main contributions are as follows:

\begin{itemize}
    \item RCAV accurately quantifies \textit{image-level} concept sensitivity both for semantically meaningful concepts and for confounding, artefactual concepts. TCAV was restricted to the \textit{dataset-level}.
    \item We highlight the \textit{false positive identification} of irrelevant concepts as a shortcoming of TCAV. RCAV uses improved hypothesis testing to significantly reduce the false positive rate.
    \item We introduce two benchmark datasets for measuring the accuracy of semantic interpretability methods in identifying concept sensitivity.
\end{itemize}

\begin{figure}
\centering
\includegraphics[width=0.9\linewidth]{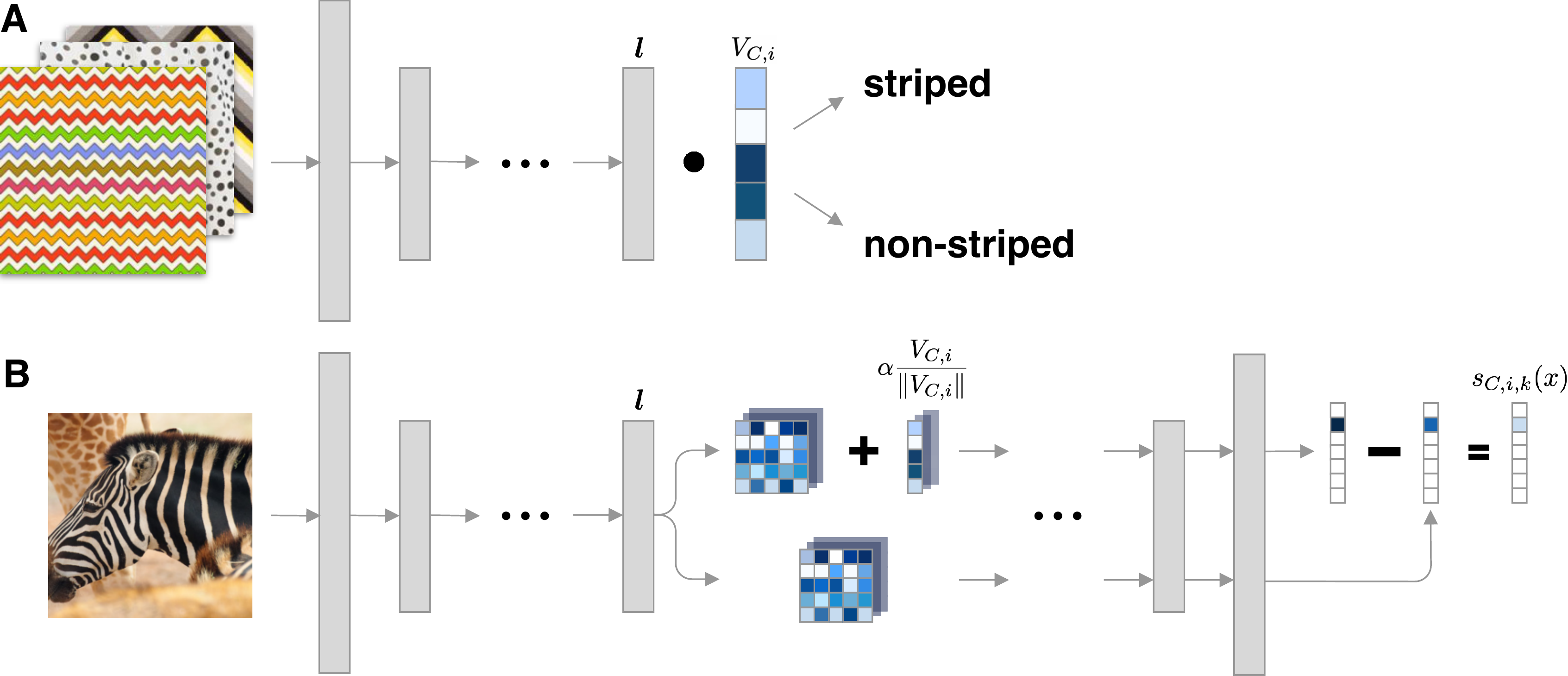}
\caption{A: RCAV defines a concept activation vector (CAV; $V_{C,i}$) by training a logistic regression on intermediate layer $l$ to distinguish the given concept from other images. B: This CAV is then used to augment inputs at test time to quantify the concept sensitivity $s_{C,i,k}(x)$ of the given input sample.}
\label{fig:diagram}
\end{figure}

\section{Related Work}

\paragraph{Motivating Semantic Interpretability} \label{motivation} It is often important to ensure that models have not inadvertently learned to predict using protected attributes and artefactual noise -- e.g. race, gender and camera blur, JPEG artefacts. Research on fairness in machine learning seeks to identify and avoid the learning of protected attributes \cite{Jiang2019-xl, Donini2018-mi, Kusner2017-db}. Previous work has relied on measuring correlation between these attributes and model predictions, but correlation is a limited proxy for the causal effect of protected attributes on prediction \cite{Chiappa2018-ua}. We suggest that semantic interpretability may prove a better metric for fairness research. Similarly, methods proposed for the de-biasing of model predictions may evaluate success using semantic interpretability methods  \cite{Li2019-ce,Kim2019-pe}. In the field of medical imaging, device-specific confounders compromise model generalization \cite{Zech2018-ad, Zech2018-qg}; these confounders may be treated as concepts and quantified using semantic interpretability methods.

\paragraph{Semantic Interpretability Methods} Our paper builds on Testing with Concept Activation Vectors (TCAV) \cite{Kim2017-yi, Zhou2018-eu}. TCAV pioneered concept activation vectors for dataset-level quantification of concept sensitivity. However, it does not support image-level concept sensitivity quantification. Adel et al. apply normalizing flows to semantic interpretability, which offers high accuracy interpretations at the cost of requiring large training sets and expensive computational overhead \cite{Adel2018-ds}. Another approach to semantic interpretability necessitates re-defining a model architecture to support interpretability, e.g. by replacing certain layers with random forests \cite{Zhang2019-wq, Agarwal2020-jt}. Architecture modification is incompatible with popular CNN architectures and so does not see widespread adoption.

\paragraph{Benchmarking Interpretability Methods} \label{relatedBench} To the best of our knowledge, no benchmarks have been published for the comparison of semantic interpretability methods. Given lack of consensus around benchmarking interpretability, recent methods have proposed re-training with feature ablation, among others \cite{Hooker2019-qg, Yang2019-rd}. Recent research has also tested the robustness of interpretability to adversarial attacks, model randomization, and input perturbation \cite{Heo2019-nw,Adebayo2018-fm,Samek2017-dw}. These existing benchmarks are not directly relevant to semantic interpretability use cases.

\begin{figure}[t]

\begin{subfigure}[t]{0.6\linewidth}
  \centering
  \includegraphics[width=.9\linewidth]{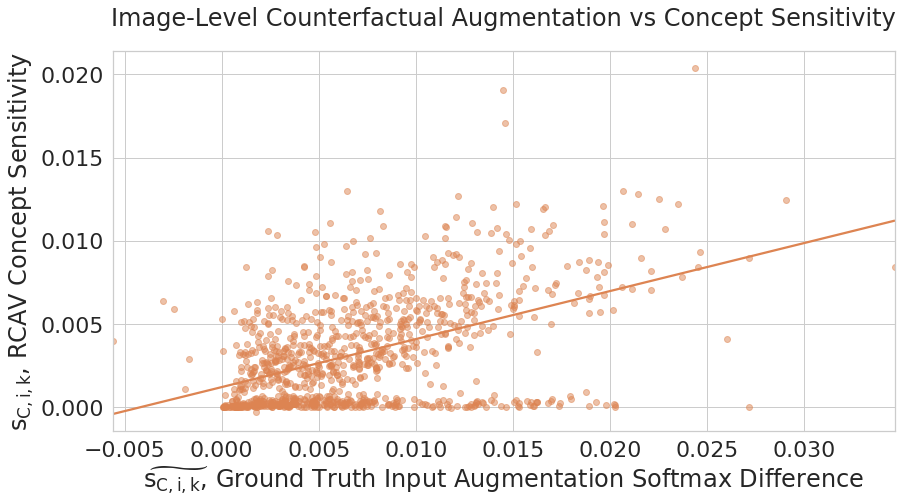}
  \caption{RCAV $s_{C,i,k}$ predicts the effect of counterfactual augmentation with Kendall's $\tau=0.31$ and p<0.05. TCAV (not shown) has $\tau=-0.06$ and p=0.77.}
  \label{fig:tfmnistcorr}
\end{subfigure} \hfill
\begin{subfigure}[t]{0.35\linewidth}
  \centering
  \includegraphics[width=0.9\linewidth]{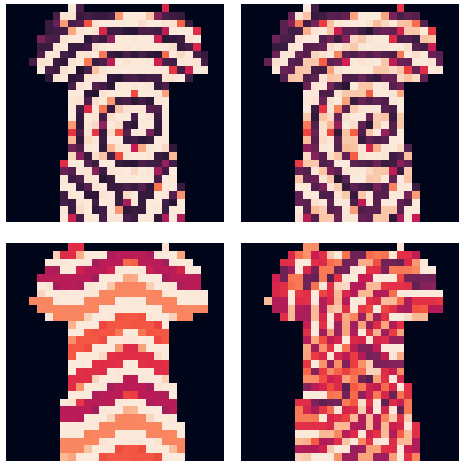}
  \caption{Samples from TFMNIST, where we interpolate between spiral and zigzag patterns in clockwise order.}
  \label{fig:tfmnistex}
\end{subfigure}
\caption{Textured Fashion MNIST provides a ground truth for semantic interpretability methods.}
 \label{fig:tfmnist}
\end{figure}

\section{Methods}

\paragraph{Overview} Semantic interpretability faces two problems: first, find a function to translate from the model's neural activations to semantic concepts; second, once we know how the model encodes a semantic concept, quantify how this concept affects model predictions -- i.e. concept sensitivity. Concept activation vectors (CAVs) address the first problem by treating a trained model as a feature extractor. Using the activations of an intermediate model layer, we train a logistic regression to identify concept samples. For example, we train a vector for each of the concepts in the set of colors red, green, blue, and yellow; the textures stripe, dot, and zigzag; or the image properties high-contrast and low-contrast. The weight vector of the logistic regression is the model's representation of our concept. To evaluate how the concept affects prediction, we augment the model's representation of inputs in the direction of the concept. \textit{No updates are needed on the model weights.}

\paragraph{Notation} We denote the trained model by $f$, the user-defined concept by $C$, and the validation set data by $X_{val}$. We denote the classes, i.e. labels, of the model's original dataset, $X_{val}$ by $k$. We denote the forward pass of a model up to and including layer $l$ by $f_l$, the forward pass of a model from the output of layer $l$ to the final layer by $f_l^+$, and the softmax score for class $k$ by $f_l^{+,k}$. We denote a concept activation vector (CAV) by $V_{C,i}$.

\subsection{Defining Concept Activation Vectors}

As in TCAV, a concept activation vector is generated by training a logistic regression on an intermediate layer $f_l$ to classify a given concept, $C_i$ relative to the union of other concepts $\bigcup\limits_{j\neq i}{C_j}$. We use 100-300 samples per concept which may be drawn from the validation set or an auxiliary dataset, as described in the appendix. The CAV,  $V_{C,i}$,  is then the weight matrix of the trained logistic regression. In effect, the CAV is a linear approximation of $f$'s encoding of the concept $C_i$.

\subsection{Quantifying Concept Sensitivity}

As shown in \autoref{fig:diagram}, we propose quantifying the sensitivity of model $f$ to the concept $C_i$, by perturbing the latent representation: \begin{equation} f_l(x) \mapsto f_l(x)+\alpha V_{C,i}/ \lVert V_{C,i} \rVert \end{equation} where $\alpha$, the step size, is a hyperparameter of RCAV shown to be insensitive within the range of 1-10 (see appendix). Then the image-level concept sensitivity score, $s_{C,i,k}(x)$, of model $f$ on input $x \in X_{val}$ to concept $C_i$ with respect to class $k$ is: \begin{equation}
s_{C,i,k}(x) = f_l^{+,k}(f_l(x)+\alpha V_{C,i}/ \lVert V_{C,i} \rVert)- f_l^{+,k}(f_l(x)) \end{equation}

The proposed RCAV sensitivity score accounts for non-linearity in $f_l^{+,k}$, i.e. the later layers of the model, but does not account for non-linearity in the initial layers of the model (this being a limitation of any CAV-based method). The score proposed in TCAV relies on the gradient's linearization of $f_l^{+,k}$, meaning TCAV linearly approximated both the initial and later layers. In \autoref{imageSense} we observe that this use of non-linearity enables RCAV to quantify the image-level concept sensitivity of $f$ whereas TCAV cannot.

At the dataset level we are also interested in determining whether the model systematically uses concept $C_i$ for prediction across all inputs of a fixed class. To that end, we compute a dataset-wide concept sensitivity score, $S_{C,i,k}$ as: \begin{equation}
S_{C,i,k} = -0.5 + \frac{1}{\lvert X_{val}^k \rvert} \sum \limits_{x \in X_{val}^k}{\mathds{1}(s_{C,i,k}(x)\geq 0)}
\end{equation}

Note the $0.5$ term centers $S_{C,i,k}$ values, so that a positive score corresponds to a positive contribution of the concept to model predictions and $0$ corresponds to concept irrelevance.

\subsection{Hypothesis Testing Robustness}

$s_{C,i,k}$ quantifies concept sensitivity, but we empirically observe that even random vectors drawn from the target space of $f_l$ yield non-zero concept sensitivity scores. We use hypothesis testing to determine whether the CAVs found by RCAV correspond to meaningful variation within the model's latent representations rather than noise. Whereas TCAV used a t-test to determine CAV significance, we find that t-testing is not robust. We instead propose a permutation test which generates a null distribution, $N$, of noise vectors, $V_{C,n}$, by permuting the correspondence between samples and labels for the concept set $C$. Then we compute a p-value: 

\begin{equation}
    p = \frac{1}{\lvert N \rvert} \sum \limits_{n \in N} {\mathds{1}(\lvert S_{C,n} \rvert \geq \lvert S_{C,i,k} \rvert)}
\end{equation}

We apply the Bonferroni correction for multiple testing to all resulting p-values. The proposed permutation test achieves a $0\%$ false positive rate on the concepts considered (\autoref{table:fpr}). In contrast to the observed $100\%$ false positive rate for the t-test used in TCAV. In \autoref{ablation}, we explain the need for a permutation test, because permutation-generated null vectors are not normally distributed, and the t-test used in TCAV underestimates the variance of the null distribution. To improve the time complexity of the proposed hypothesis test, we stop computing the p-value as soon as a null result is guaranteed. RCAV, using a significance threshold of $0.05$ and $500$ permutations, achieves a best case 250x speed-up and a 4x average case improvement compared to TCAV.

RCAV leaves layer, concept set, and step size as hyperparameters to be determined by the user; these choices may be tuned to minimize variance of the permutation null set and maximize CAV accuracy. To comprehensively assess RCAV independently of manual layer choice, we select five layers uniformly spaced across model depth. We test on a subset of layers instead of all layers, because the worst case runtime of RCAV scales quadratically with the number of layers tested. Since the permutation test is an approximate method, the number of permutations must increase linearly against number of layers to avoid losing power following multiple testing adjustment. Concept set images are taken from the validation set. For example, the concept red, is represented by the 100 validation set images with highest intensity in the red channel. We discuss choice of concept set and step size in detail in the appendix. In the below experiments, we use $\alpha=10$ throughout.

\section{Benchmarking Semantic Interpretability Methods}

\begin{figure}[t]
\centering
\begin{subfigure}[t]{0.6\linewidth}
  \centering
  \includegraphics[width=.9\linewidth]{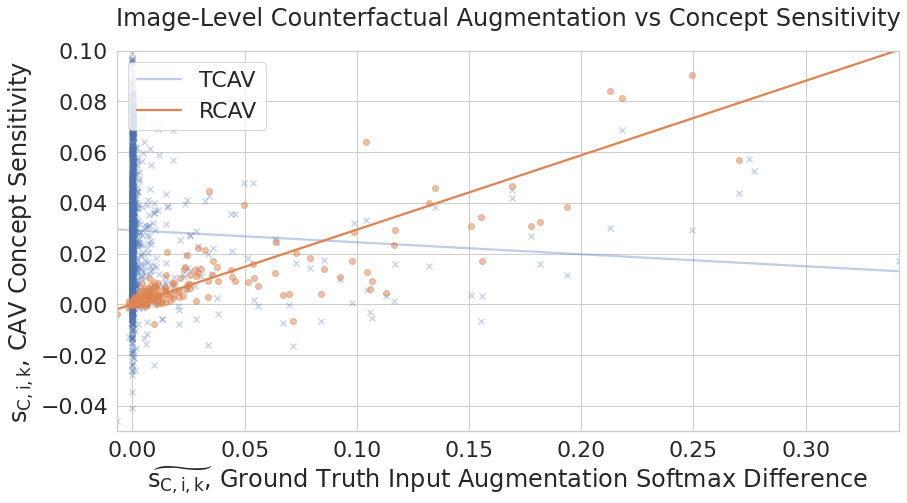}
  \caption{RCAV $s_{C,i,k}$ predicts counterfactual augmentation with Kendall's $\tau=0.85$ and p<0.05. TCAV has $\tau=-0.09$ and $p=0.42$. Positive and negative axes inverted for clarity.}
  \label{fig:camelyoncorr}
\end{subfigure} \hfill
\begin{subfigure}[t]{0.35\linewidth}
  \centering
  \includegraphics[width=0.9\linewidth]{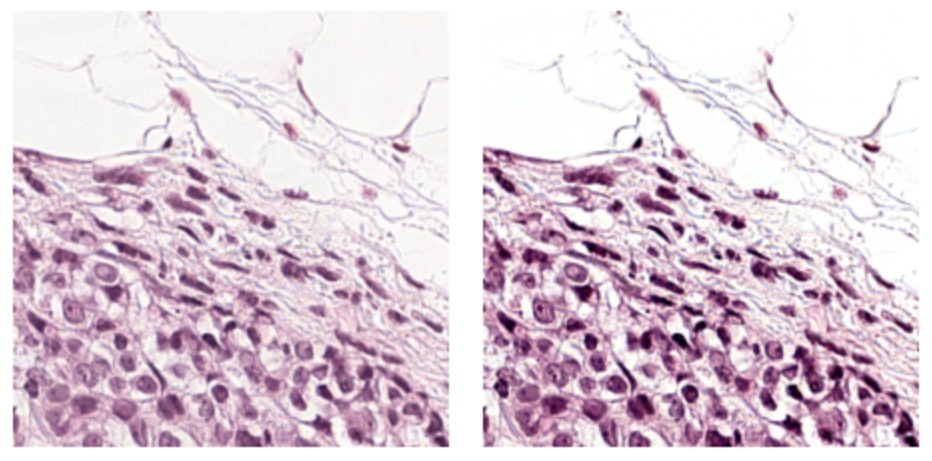}
  \caption{Samples from CAMELYON16 with default and augmented contrast levels.}
  \label{fig:camelyonex}
\end{subfigure}
\caption{Biased contrast-augmented CAMELYON16 dataset provides a ground truth for assessing how accurate semantic interpretability methods are at detecting bias. }
 \label{fig:camelyon}
\end{figure}

To evaluate RCAV's performance, we introduce two new datasets that test the accuracy of concept sensitivity measurements. We designed these datasets to come as close as possible to real world use of semantic interpretability methods. Existing benchmarks evaluate interpretability methods without taking into account the intended use case. For example, Lage et al. quantify interpretability by measuring how accurately users can reconstruct model predictions from interpretability explanations \cite{Lage2018-px}. Although this reconstruction metric is important, it does not tell us how reliably the interpretability method will work in practice. 

Interpretability methods are used to determine whether a model relies on robust input features, or instead relies on spurious noise. To determine the ground truth about the relative importance of signal and noise, we need a dataset in which we have two copies of each image: one with the concept and one without. For example, for the image contrast concept, we would run the model on an image with normal contrast levels, and then we would run the model on the \textit{same image} with increased contrast levels. The difference in prediction after augmentation gives us a ground truth regarding how sensitive the model is to contrast in that image.

\subsection{Benchmarking Datasets}
\label{benchmarks}

\paragraph{Textured Fashion MNIST (TFMNIST)}
\label{TFMNIST}

To evaluate semantic interpretability we need a dataset where we can manipulate the presence of concepts as realistically as possible. Previous work superimposed foreground images on background, but this process results in unrealistic images such as a backpack floating on a bamboo forest \cite{Yang2019-rd}. We build on the Fashion MNIST (FMNIST) dataset \cite{Xiao2017-ue} by replacing the surface of clothing items with our concepts. We use textures drawn from a Google Image search to replace the original surfaces as shown in \autoref{fig:tfmnist}. Importantly, our dataset is fully extensible: users may easily replace these textured concepts with any concept of their choosing -- colors, animals, etc. 

Using FMNIST, we may continuously interpolate between two arbitrary textures creating, for example, shirts with mixtures of striped and spiral patterns. We construct a training set in which all T-shirts are given spiral textures and all non-T shirts are given zigzag textures. All other classes are given dotted and striped textures. Then, on the validation set, we interpolate between textures to compute the ground truth concept sensitivity. For example, by interpolating spiral T-shirts 10\% in the direction of zigzag textures, we can quantify the effect of zigzags on t-shirt predictions. See top row of \autoref{fig:tfmnistex}.

\paragraph{Biased CAMELYON16}

For application to medical imaging, we need a dataset where we can manipulate the presence of confounding artefacts. To simulate device differences, we build on the CAMELYON16 lymph node section histology dataset \cite{Bejnordi2016-mn} by augmenting the contrast level of images. Importantly, this dataset is fully extensible: contrast augmentation may be replaced by any other form of corruption -- JPEG artefacting, camera tilt, out-of-focus, etc. 

For our experiments, we increased the contrast of cancerous tissue while leaving non-cancerous tissue at baseline, \autoref{fig:camelyonex}. Then, to provide a ground-truth for the model's sensitivity to the contrast concept, we compare model predictions before and after changing contrast level by 3\% -- a change nearly imperceptible to the human eye.

\paragraph{ImageNet}

We can only rely on the results of a semantic interpretability method if we know that the method will not report high concept sensitivity where there is none. To this end, we use ImageNet to quantify the robustness of semantic interpretability methods to false positives \cite{Deng2009-dm}. We build on the experiments proposed in \cite{Kim2017-yi} for texture and color concepts. We select negative control classes which appear intuitively unrelated to the concepts: great white shark for texture, and apron for color (the first unrelated class by ImageNet label order). This protocol for identifying negative control classes may be generally applied to any multi-class classification task.

\begin{center}
    \begin{table}[t]
      \caption{Interpretability performance metrics. AUROC and AUPRC after applying 75th percentile threshold to label-binarize counterfactual augmentation differences.}
      \label{table:metrics}
      \centering
      \begin{tabular}{llllllll}
        \toprule
        \multicolumn{2}{c}{Dataset:} & \multicolumn{3}{c}{Textured FMNIST} & \multicolumn{3}{c}{Biased CAMELYON16}   \\
        \midrule
        \multicolumn{2}{c}{Metric:} & $P_\tau$ & AUROC & AUPRC &    $P_\tau$ & AUROC & AUPRC \\
        \midrule
        \multicolumn{2}{c}{RCAV} & \textbf{64\%} & \textbf{0.77} & \textbf{0.56}& 
            \textbf{92\%} & \textbf{0.94} & \textbf{0.89}  \\
        \multicolumn{2}{c}{TCAV} & 53\% & 0.51 & 0.24 & 
            45\% & 0.55 & 0.27  \\
        \bottomrule
      \end{tabular}
    \end{table}
\end{center}

\subsection{Measuring Concept Sensitivity}
\label{metrics}

When evaluating a semantic interpretability method on the proposed datasets, we highlight the need for metrics which align with the intended use case. Interpretability methods are used at test time to decide whether the model is trustworthy. The user decides whether to accept the model's prediction as right for the right reasons, or to reject the model's prediction as unjustified. Given a ground truth for the model's counterfactual sensitivity to a concept, we compute the area under the receiver operating characteristic curve (AUROC) and area under the precision-recall curve (AUPRC). The proposed datasets provide a ground truth for concept sensitivity as the difference between the model's predictions before and after input augmentation, denoted $\widetilde{s_{C,i,k}}$. For example in CAMELYON16, we augment the input images by reducing the contrast level, and then depending on whether model prediction delta exceeds a certain threshold, these samples are labelled "trustworthy" or "untrustworthy."

AUROC and AUPRC metrics are typically used to quantify performance for tasks with fixed decision thresholds. However, in some cases the interpretability method will be used on-the-fly without a fixed decision threshold. For example, a user may be interested in identifying ImageNet samples in a batch which are most sensitive to the stripes concept. To evaluate accuracy for this use case, we compute the probability, $P_\tau$, of RCAV correctly ordering the image-level concept sensitivity for each pair of samples. We denote the ground truth model concept sensitivity by $\widetilde{s_{C,i,k}}$, and validation set size by $n$. 

\begin{equation}
    P_\tau = \dfrac{1}{\binom{n}{2}}\sum\limits_{x_i,x_j \in X_{val}}{\mathds{1}(s_{C,i,k}(x_i)\geq s_{C,i,k}(x_j) \equiv \widetilde{s_{C,i,k}(x_i)}\geq \widetilde{s_{C,i,k}(x_j)})} 
\end{equation}

When reporting $P_\tau$ and Kendall's $\tau$ values, we suggest generating p-values by permutation testing and comparing observed $\tau$ against permutation test-generated $\tau$. The above metrics are applicable to other datasets, semantic interpretability methods, and even saliency maps.

\section{Results}
\label{results}

The proposed datasets define ground truth concept sensitivity that we now use to evaluate the accuracy of RCAV. For the FMNIST and CAMELYON16 datasets, we use Inception-v3 \cite{Szegedy2015-xz}, and for FMNIST in particular, we apply input mixup during training \cite{Zhang2017-oh}. Input mixup has been shown to locally linearly regularize out-of-distribution predictions \cite{Guo2019-of}. For the ImageNet experiments we use GoogLeNet to allow direct comparison against TCAV results \cite{Szegedy2014-wg}.

\subsection{Image-level Concept Sensitivity}
\label{imageSense}

By comparing RCAV predicted concept sensitivity, $s_{C,i,k}$, to ground truth concept sensitivity, $\widetilde{s_{C,i,k}}$, \autoref{fig:tfmnist} and \autoref{fig:camelyon} show that RCAV accurately predicts the concept sensitivity of individual model predictions. The correlation between $s_{C,i,k}$ and $\widetilde{s_{C,i,k}}$ is significant following permutation testing. The RCAV linear fit has intercept at or near 0, indicating that RCAV concept sensitivity predictions are calibrated such that a null sensitivity result in RCAV indicates ground truth null and vice versa. 

Unlike RCAV, TCAV concept sensitivity correlation with ground truth is indistinguishable from the null (\autoref{fig:camelyon}), meaning TCAV explanations perform no better than random. The failure of TCAV to predict observed concept sensitivity suggests that even for the small perturbations considered, the first-order, gradient approximation used by TCAV fails.

\autoref{table:metrics} shows that RCAV performs robustly across choice of metric. For CAMELYON16, RCAV's image-level sensitivity performance is near optimal, but for FMNIST we see a drop. In \autoref{limitations}, we explain the poor FMNIST performance relative to Biased CAMELYON16 by performing a singular value decomposition (SVD) component analysis to place a bound on RCAV performance.

\subsection{Dataset-level False Positive and False Negative Robustness}

\begin{center}
    \begin{table}[t]
      \caption{False positive rate for hypothesis testing on negative controls.}
      \label{table:fpr}
      \centering
      \begin{tabular}{llllllll}
        \toprule
        \multicolumn{2}{c}{Concept and Class:} & \multicolumn{2}{c}{Color for Apron} & \multicolumn{2}{c}{Texture for Shark}   \\
        \midrule
        \multicolumn{2}{c}{Multiple Testing Adjustment:} & Before & After & Before & After\\
        \midrule
        \multicolumn{2}{c}{RCAV}    & \textbf{5  \%} & \textbf{0  \%}    & \textbf{13  \%}& \textbf{0  \%}   \\
        \multicolumn{2}{c}{TCAV}    & 100  \% & 100  \%  & 100  \% & 100  \%   \\
        \bottomrule
      \end{tabular}
    \end{table}
\end{center}

\begin{figure}[h]
\centering
    \begin{subfigure}[t]{0.49\linewidth}
      \centering
      \includegraphics[width=1\linewidth]{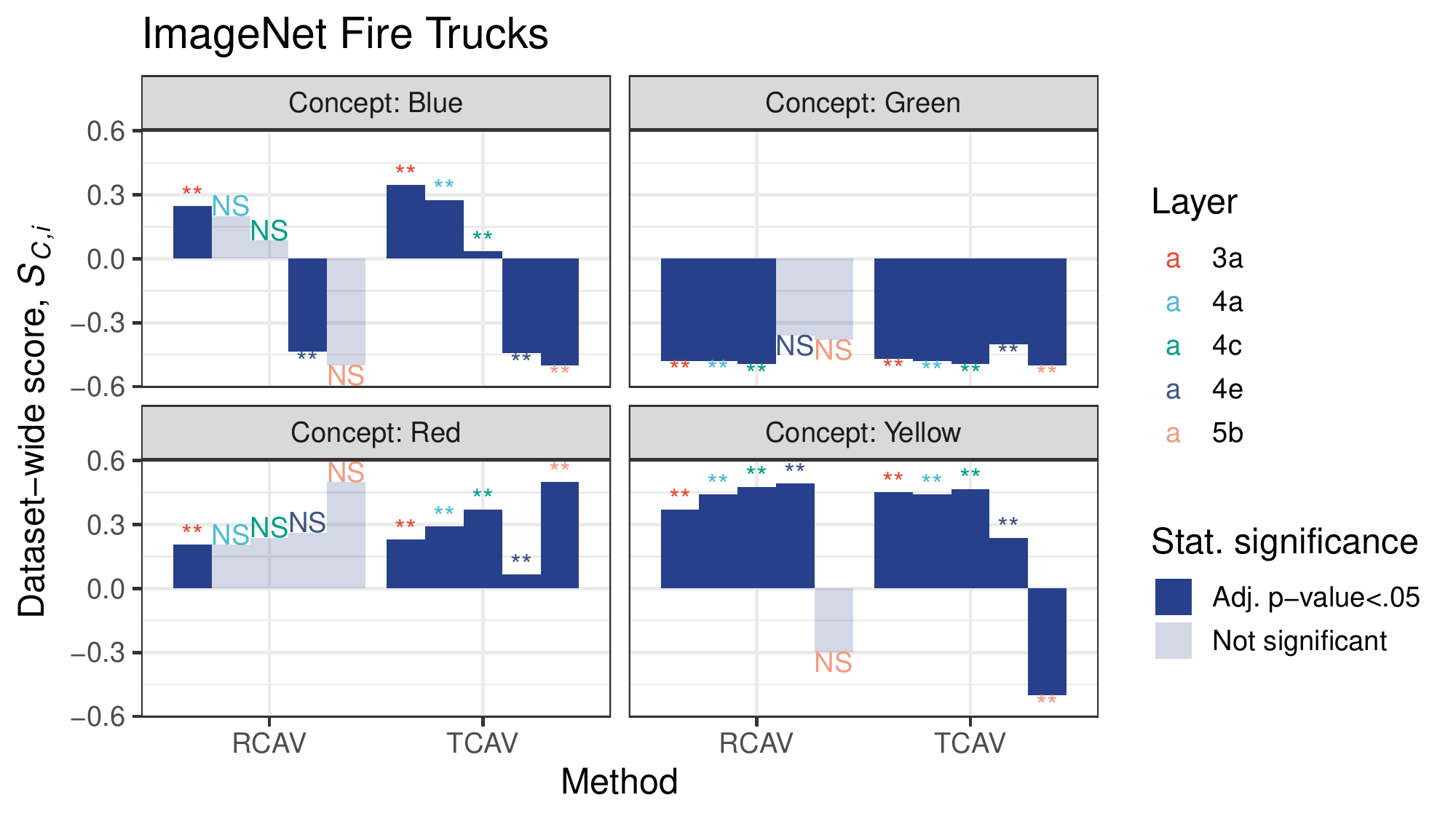}
      \caption{}
      \label{fig:firetruck}
    \end{subfigure} \hfill
    \begin{subfigure}[t]{0.49\linewidth}
      \centering
      \includegraphics[width=1\linewidth]{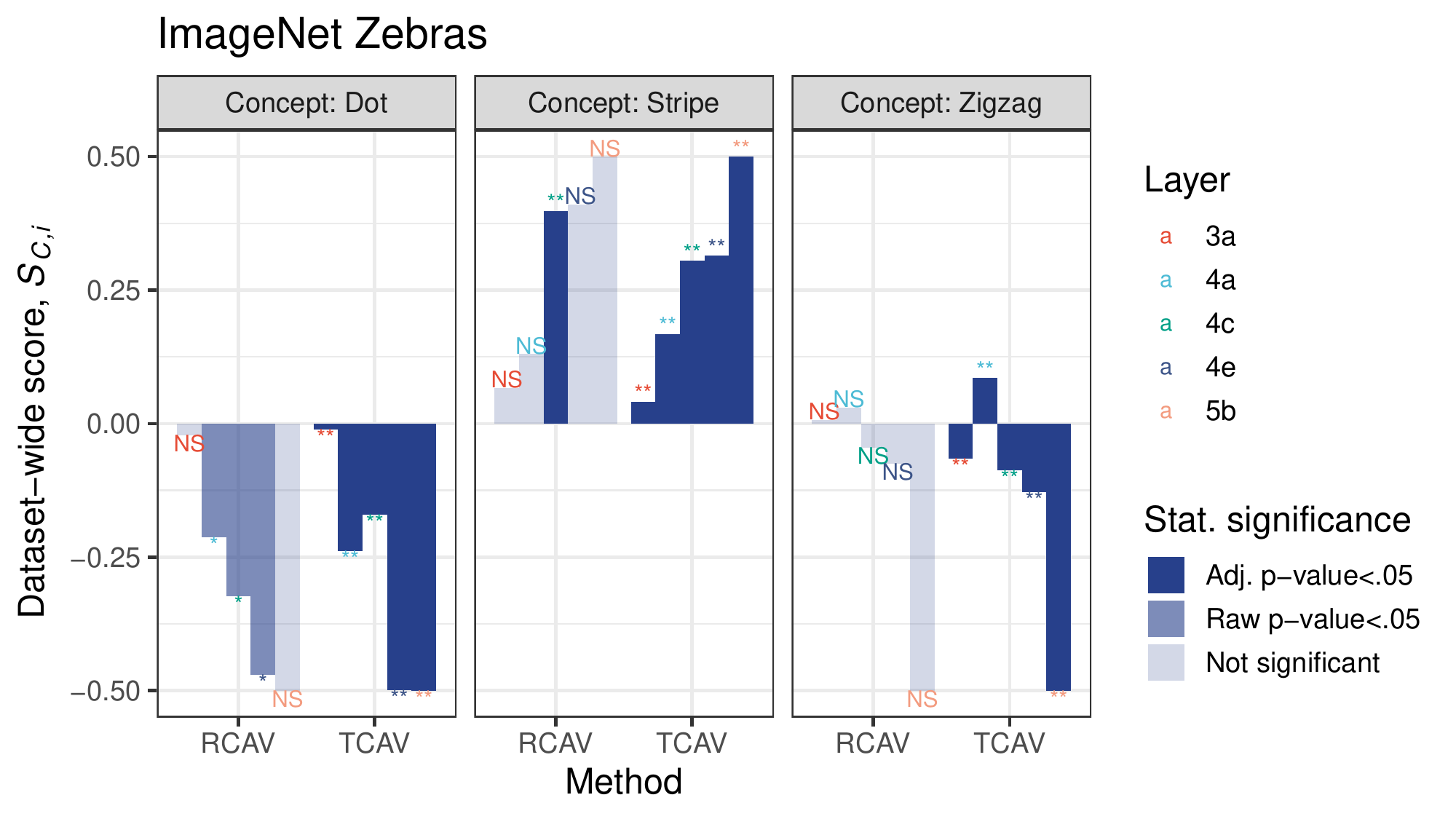}
      \caption{}
      \label{fig:zebra}
    \end{subfigure}
    \caption{RCAV and TCAV quantify dataset-level concept sensitivity in ImageNet using GoogLeNet.}
     \label{fig:imagenet}
\end{figure}

To quantify RCAV's robustness to false positives, we select negative control ImageNet classes which appear unrelated to the texture and color concepts. We compute the p-values of $S_{C,i,k}$ for these classes. When the p-values pass the significance threshold, we call this a false positive. As would be the case in real-world applications of RCAV, we do not have access to the ground truth regarding model sensitivity to color on aprons and to texture on great white sharks, but we have selected these pairs in order to maximize the apparent likelihood of non-interaction.\footnote{Alternatively, we may quantify the false positive rate by applying RCAV to an untrained model. In this case, the ground truth is known. However, an untrained model often yields lower CAV accuracy, and so it is unclear whether a low false positive rate for untrained models generalizes to the trained model setting.} In \autoref{table:fpr}, we see that RCAV has a 0\% false positive rate for both concepts considered, whereas TCAV predicts a false positive on every layer tested. In \autoref{ablation}, we clarify which aspects of the RCAV method contribute to this decrease in false positives. 

Since RCAV uses more stringent hypothesis testing, we reconstruct two experiments proposed in the TCAV paper to demonstrate that RCAV does not suffer increased false negatives, see \autoref{fig:imagenet}. Significance testing in TCAV is not meaningful because all results are found to be significant (dark blue bars), whereas RCAV finds layers in which concepts are most meaningfully encoded versus those that are not (light gray-blue bars). There is no known ground truth for \autoref{fig:imagenet}, so it remains open whether, for instance, GoogLeNet uses the concept zigzag for making zebra predictions.

\subsection{Ablation Study}
\label{ablation}

In order to clarify the extent to which each aspect of RCAV is responsible for the observed reduction in false positives, we run an ablation study in which we hold constant the sensitivity scoring functions used in RCAV and TCAV -- softmax difference and cosine similarity, respectively -- while varying the hypothesis test. RCAV improves over TCAV by replacing cosine similarity with softmax difference, and by replacing a t-test with a permutation test. As an intermediate alternative to the t-test and permutation tests, we also include a null hypothesis defined by vectors drawn from a unit norm uniform distribution. The uniform null p-value is calculated using the same formula as the permutation test p-value (equation 4). We report the raw p-value false positive rate before multiple testing adjustment, because raw p-values give a worst-case upper bound on the observed FPR independent of layer choice. \autoref{table:ablation} shows that permutation testing is necessary to reduce the false positive rate, but the sensitivity scoring function appears to also help. We suggest that future use of hypothesis testing in interpretability favor permutation testing where possible over parametric tests.

\begin{center}
    \begin{table}[t]
      \caption{Worst-case analysis of false positive rate by choice of hypothesis test. Texture concept for class great white sharks. FPR calculated using non-adjusted p-values. Default settings proposed by RCAV and TCAV in italics, best result in bold.}
      \label{table:ablation}
      \centering
      \begin{tabular}{llllllll}
        \toprule
                            & RCAV      & TCAV      \\
        \midrule
        Permutation test    & \textit{\textbf{13 \%}}     & 60 \%     \\
        Uniform random null & 40 \%     & 80 \%     \\
        t-test              & 93 \%     & \textit{100 \%}    \\
        \bottomrule
      \end{tabular}
    \end{table}
\end{center}

\section{Conclusion}

In this paper, we propose a novel use of CAVs as well as novel benchmarks for the evaluation of semantic interpretability methods. Our RCAV method is designed to be as user-friendly as saliency maps by supporting on-the-fly explanations for individual model predictions. At the dataset level, RCAV enables the quantification of model biases into a single scalar value. We also address a gap in the interpretability benchmarking literature by proposing datasets and metrics for the realistic evaluation of semantic interpretability methods. On these benchmarks, we demonstrate that RCAV improves upon previous work in terms of accuracy, robustness, and runtime.

\paragraph{Limitations} \label{limitations} Although we observe results to be robust across choice of hyper-parameters in most cases, on the FMNIST dataset, image-level concept sensitivity is best captured on layers 6a and 6b with null results on other layers. An SVD analysis of latent space pairwise differences before and after input augmentation showed that these are the layers with highest explained variance in the first component (see appendix). This SVD analysis implies that for the rest of the layers, texture concepts are not encoded uniformly across the validation set, so CAV-based methods are ineffective.

\paragraph{Future Work} Recent research has used CAVs to automatically discover prediction-relevant concepts \cite{Ghorbani2019-ki}. RCAV may enhance such methods, serving as a drop-in replacement for TCAV. This paper shows that RCAV improves over TCAV across benchmarks, but in the unlimited data setting, we expect non-linear methods to outperform CAV methods. A complete comparison of RCAV to other semantic interpretability methods remains an outstanding direction for future research.

\section*{Broader Impact}

RCAV is broadly applicable to the real world deployment of CNNs. We hope that RCAV will prove of use to researchers and engineers as a technique to assist in guaranteeing the robustness and fairness of models. We also see potential applications to scientific discovery: scientific properties can be used to define concept sets and RCAV can be applied to assess the scientific significance of certain properties. For example, it may be possible to train a model on molecular similarity and then assess the importance of molecule properties using RCAV. We hope that the metrics and benchmarks proposed above will encourage greater reproducibility and transparency around the development of semantic interpretability methods. In particular, the proposed false positive metric will ideally help avoid mistaken application of semantic interpretability methods to the high-stakes fields of medical imaging.

\begin{ack}
We would like to thank Laura Gunsalus and Garret Gaskins for extensive feedback on a draft of this paper. We would also like to thank Wren Saylor, Will Connell and Kangway Chuang for feedback on an early draft. This work was in part supported by the Helen Diller Family Comprehensive Cancer Center Impact Award and the Melanoma Research Alliance. 
\end{ack}

\printbibliography

\pagebreak 

\appendix

\section{Choice of Concept Set}

For RCAV, a set of representative images defines a concept. For any semantic concept there will be many possible choices of concept set: e.g. we may choose between either color patches or colored objects to define a particular color. In practice, it is usually easiest to draw concept set images from the validation set. For instance, to define the concept red, we could select the 100 validation set images with highest intensity in the red channel. When sampling images for the concept set, it is necessary to maintain class balance -- i.e. the number of samples per class for each sub-concept must be identical. For the TFMNIST and Camelyon experiments we use validation set images to define the concept sets. In contrast, for the ImageNet experiments, we follow the precedent set in TCAV using Broden images for textures and Gaussian-noised color patches for color \cite{Kim2017-yi, Bau2017-yo}.

If there are multiple possible choices of concept set, the optimal choice minimizes the variance of the set of null concept sensitivity scores, $\{S_{C,n}\}_{n \in N}$. This optimal choice of concept set will minimize false negatives, i.e. misidentification of concepts that are meaningful to model prediction as statistically insignificant.

\section{Hyperparameter Sensitivity Analysis}

\begin{center}
    \begin{table}[h]
      \caption{Performance as a function of layer for contrast-augmented CAMELYON16.}
      \label{table:layer}
      \centering
      \begin{tabular}{llllllll}
        \toprule
        \multicolumn{2}{c}{} & \multicolumn{3}{c}{RCAV} & \multicolumn{3}{c}{TCAV}   \\
        \midrule
        \multicolumn{2}{c}{} & $P_\tau$ & AUROC & AUPRC &    $P_\tau$ & AUROC & AUPRC \\
        \midrule
        \multicolumn{2}{c}{Conv2d 3b 1x1} & 92\% & 0.94 & 0.89 & 
            45\% & 0.55 & 0.27  \\
        \multicolumn{2}{c}{Mixed 5c}     & 89\% & 0.94 & 0.86 & 
            68\% & 0.65 & 0.31  \\
        \multicolumn{2}{c}{Mixed 6d}     & 83\% & 0.84 & 0.77 & 
            69\% & 0.70 & 0.34  \\
        \multicolumn{2}{c}{Mixed 7b}     & 91\% & 0.94 & 0.83 & 
            70\% & 0.70 & 0.37  \\
        \bottomrule
      \end{tabular}
    \end{table}
\end{center}

\paragraph{Layer} Interpretability methods seek to explain model predictions. We propose using layer-specific RCAV scores for interpretability, but we may only extrapolate from layer-specific results if RCAV performance is layer invariant. At the image level, \autoref{table:layer} shows that RCAV predicts concept sensitivity for all layers considered. 

At the dataset level, we observe that the absolute value of $S_{C,i,k}$ increases monotonically as the layer approaches softmax. We explain this increase by observing that head of the model, $f^+_l$, converges to linearity as $l$ approaches $l_{max}$. In the linear case, any fixed CAV will have $\lvert S_{C,i,k} \rvert = 0.5$, because the effect of perturbation in a fixed direction is invariant over choice of input for a linear classifier. 

To ensure that any conclusion based on a specific layer's RCAV scores is representative of the whole model, we need a weak form of layer invariance: consistency of $\text{sign}(S_{C,i,k})$ across layers. \autoref{fig:imagenet} shows that for six out of the seven concepts considered, RCAV consistently predicts dataset-level concept sensitivity. It is possible that layer inconsistency occurs when a concept plays a non-binary role in the classifier's decision function. For this reason, we recommend testing multiple layers when using RCAV for dataset-level concept sensitivity quantification.

\begin{center}
    \begin{table}[h]
      \caption{Performance as a function of step size for contrast-augmented CAMELYON16, using layer Conv2d3b.}
      \label{table:stepsize}
      \centering
      \begin{tabular}{llllllll}
        \toprule
        \multicolumn{1}{c}{Step Size} & $AVG(s_{C,i,k})$ & $P_\tau$ & AUROC & AUPRC \\
        \midrule
        \multicolumn{1}{c}{0.1} & 2e-5 & 81\% & 0.93 & 0.88 \\
        \multicolumn{1}{c}{1}   & 2e-4 & 88\% & 0.94 & 0.86 \\
        \multicolumn{1}{c}{10}  & 2e-3 & 94\% & 0.94 & 0.89 \\
        \multicolumn{1}{c}{100} & 0.03 & 91\% & 0.94 & 0.86 \\
        \bottomrule
      \end{tabular}
    \end{table}
\end{center}

\paragraph{Step size} In real-world use of RCAV, ground truth concept sensitivity is not known, so it is impossible to tune step size for optimal performance. Instead we suggest choosing step size such that observed concept sensitivity scores, $s_{C,i,k}$, range from 0.001 to 0.1. This is the observed range of softmax differences in the benchmark experiments shown in \autoref{fig:tfmnist} and \autoref{fig:camelyon}. Empirically, RCAV performance is robust across choice of step size, as shown in \autoref{table:stepsize}.

\begin{center}
    \begin{table}[h]
      \caption{Performance as a function of label binarization threshold for contrast-augmented CAMELYON16, using layer Conv2d3b.}
      \label{table:threshold}
        \centering
        \begin{tabular}{llllllll}
        \toprule
      \multicolumn{1}{c}{} & \multicolumn{2}{c}{RCAV} & \multicolumn{2}{c}{TCAV}   \\
        \midrule
        \multicolumn{1}{c}{Threshold} & AUROC & AUPRC & AUROC & AUPRC \\
        \midrule
        \multicolumn{1}{c}{5\%}  & 0.94 & 0.99 & 0.35 & 0.84\\
        \multicolumn{1}{c}{25\%} & 0.97 & 0.99 & 0.38 & 0.68\\
        \multicolumn{1}{c}{75\%} & 0.94 & 0.89 & 0.55 & 0.27\\
        \multicolumn{1}{c}{95\%} & 0.98 & 0.90 & 0.63 & 0.07\\
        \bottomrule
      \end{tabular}
    \end{table}
\end{center}

\paragraph{Label binarization threshold} In practice, we often use RCAV to make a binary decision: either the input is sensitive to the concept, or the input is not sensitive to the concept. In \autoref{metrics}, we used AUROC and AUPRC to quantify the accuracy of RCAV for this binary task. The ground truth for this task is whether model prediction delta exceeds a certain threshold when augmenting inputs. Formally the ground truth labels are defined by $x \mapsto \mathds{1}(f^k(x)-f^k(x')>t)$ for some fixed threshold $t$, augmented input $x'$ and class $k$. In \autoref{table:threshold} we choose our threshold as a percentile of the ground truth sensitivity values, and show that RCAV performs robustly across all thresholds.

\section{Measuring Concept Encoding Linearity}

\begin{center}
    \begin{table}[h]
      \caption{Accuracy of rank one approximations to concept latent encodings.}
      \label{table:svd}
      \centering
      \begin{tabular}{llllllll}
        \toprule
        \multicolumn{2}{c}{Layer} & TFMNIST & CAMELYON16 \\
        \midrule
        \multicolumn{2}{c}{Conv2d 3b 1x1} & 24\% & 50\% \\
        \multicolumn{2}{c}{Mixed 5c}      & 40\% & 37\%  \\
        \multicolumn{2}{c}{Mixed 6d}      & 40\% & 80\% \\
        \multicolumn{2}{c}{Mixed 7b}      & 44\% & 84\%  \\
        \bottomrule
      \end{tabular}
    \end{table}
\end{center}

RCAV relies on CAV's linear approximation of the model's concept encoding. By doing an SVD on the ground truth concept sensitivity differences, we can measure the extent to which this linearity constraint bottlenecks RCAV performance. The CAV can reliably estimate the ground truth effect only if the difference vector between encodings of input, $x$, and augmented input, $x'$, is similar to the CAV -- i.e. $\lVert V_{C,i} -  (f_l(x)-f_l(x'))\rVert<\epsilon$. If, on the other hand, the difference vector has high variance across points of the validation set, then the effect of the concept cannot be encoded as a CAV. We can measure the extent to which the concept is consistently encoded by examining the matrix of pairwise encoding differences, 
\begin{equation}
    D_l = \begin{bmatrix} 
        f_l(x_0)-f_l(x_0') \\
        \vdots  \\
        f_l(x_n)-f_l(x_n')
        \end{bmatrix}
\end{equation}

The optimal CAV\footnote{In practice, the optimal CAV cannot be calculated in this way, because it is not feasible to counterfactually augment the input -- i.e. we do not have $x'$.} is the first singular vector for $D_l$, because this vector best approximates $f_l(x_i)-f_l(x_i')$. Using the SVD, we can upper bound the performance of RCAV on layer $l$ by calculating the reconstruction accuracy, $r$, of the best rank one approximation to $D_l$. Matrix dimensions varies across layer, so we normalize the reconstruction accuracy to $r=\lVert D_l-D_{l,1} \rvert / \lVert D_l \rVert$ where $D_{l,1}$ is the rank one approximation and we use the Frobenius norm. \autoref{table:svd} shows that reconstruction accuracy is higher for CAMELYON16 than TFMNIST. We infer that the CAMELYON16 model's encoding of the contrast concept is more linear than the TFMNIST model's encoding of the texture concepts. These results explain the difference in performance between these two datasets seen in \autoref{table:metrics}.

\end{document}